\documentclass[runningheads]{llncs}

\usepackage{eccv}

\usepackage{eccvabbrv}

\usepackage{graphicx}
\usepackage{booktabs}

\usepackage{multirow}

\usepackage[accsupp]{axessibility}  % Improves PDF readability for those with disabilities.

\usepackage{hyperref}

\usepackage{orcidlink}

\begin{document}

\title{Delayed Bidirectional Alignment via Disentangled Audio Semantics for Audio-Visual Segmentation} 

\titlerunning{DDAVS}

\author{Jingqi Tian\inst{1} \and
Yiheng Du\inst{2} \and
Haoji Zhang\inst{1} \and
Yuji Wang\inst{1} \and
Isaac Ning Lee\inst{1} \and \\ 
Xulong Bai\inst{1} \and 
Tianrui Zhu\inst{1} \and
Jingxuan Niu\inst{1} \and
Yansong Tang\inst{1}$^\dagger$}

\authorrunning{J.~Tian et al.}

\institute{
Tsinghua Shenzhen International Graduate School, Tsinghua University, China \and
Peking University, China \\
 \email{\{tjq25@mails, tang.yansong@sz\}.tsinghua.edu.cn}
}

\maketitle

\begingroup
\renewcommand{\thefootnote}{\ensuremath{\dagger}}
\footnotetext{Corresponding author}
\endgroup
\setcounter{footnote}{0}

\begin{abstract}
Audio–Visual Segmentation (AVS) aims to localize sound-producing objects at the pixel level by integrating auditory and visual cues. However, existing methods often struggle with multi-source entanglement and audio–visual misalignment, leading to a dominance bias toward acoustically or visually salient objects (i.e., louder or larger ones) at the expense of subtler or co-occurring sources.
To address these challenges, we propose DDAVS: Delayed Bidirectional Alignment via Disentangled Audio Semantics for Audio-Visual Segmentation. To mitigate multi-source entanglement, DDAVS employs learnable queries to extract audio semantics and anchor them within a structured semantic space derived from an audio prototype memory bank. This process is further optimized through contrastive learning to enhance discriminability and robustness. To alleviate audio–visual misalignment, DDAVS introduces dual cross-attention with delayed modality interaction, improving the robustness of multimodal alignment.
Extensive experiments on the AVS-Objects and VPO benchmarks demonstrate that DDAVS achieves state-of-the-art performance across single-source, multi-source, and multi-class multi-instance scenarios. These results validate the effectiveness and generalization ability of our framework under challenging real-world audio–visual segmentation conditions. \href{https://trilarflagz.github.io/DDAVS-page/}{Project page.} 
% Extensive experiments on the AVS-Objects and VPO benchmarks demonstrate that DDAVS achieves state-of-the-art performance, consistently outperforming existing methods across single-source, multi-source, and multi-class multi-instance scenarios. These results validate the effectiveness and generalization ability of our framework under challenging real-world audio–visual segmentation conditions.
\end{abstract}

\section{Introduction}
\label{sec:intro}
Traditional Visual Segmentation (VS) focuses solely on appearance, 
partitioning all visible objects in an image regardless of their physical state or behavior \cite{kirillov2023segment, ke2023segment}.
In contrast, Audio–Visual Segmentation (AVS) \cite{liu2023audio, liu2025dynamic} introduces an additional auditory modality, aiming to identify and segment \textit{sound-emitting} objects that are temporally and semantically linked to the accompanying audio signal. 
By enforcing pixel-level alignment between auditory cues and visual evidence, 
AVS moves toward a more holistic understanding of acoustic and visual multi-modal scenes, alongside recent advances in fine-grained grounding, temporal consistency and unified reasoning
\cite{damonlpsg2025videollama3, jin2025videomem, sun2024video, zhu2025memorize, jin2026dgpo, du2025crab, ma2026safe, jin2025videocurl, lin2026visd, wang2024refavs}.

\begin{figure}[t]
    \centering
    \includegraphics[width=1\textwidth, keepaspectratio=true]{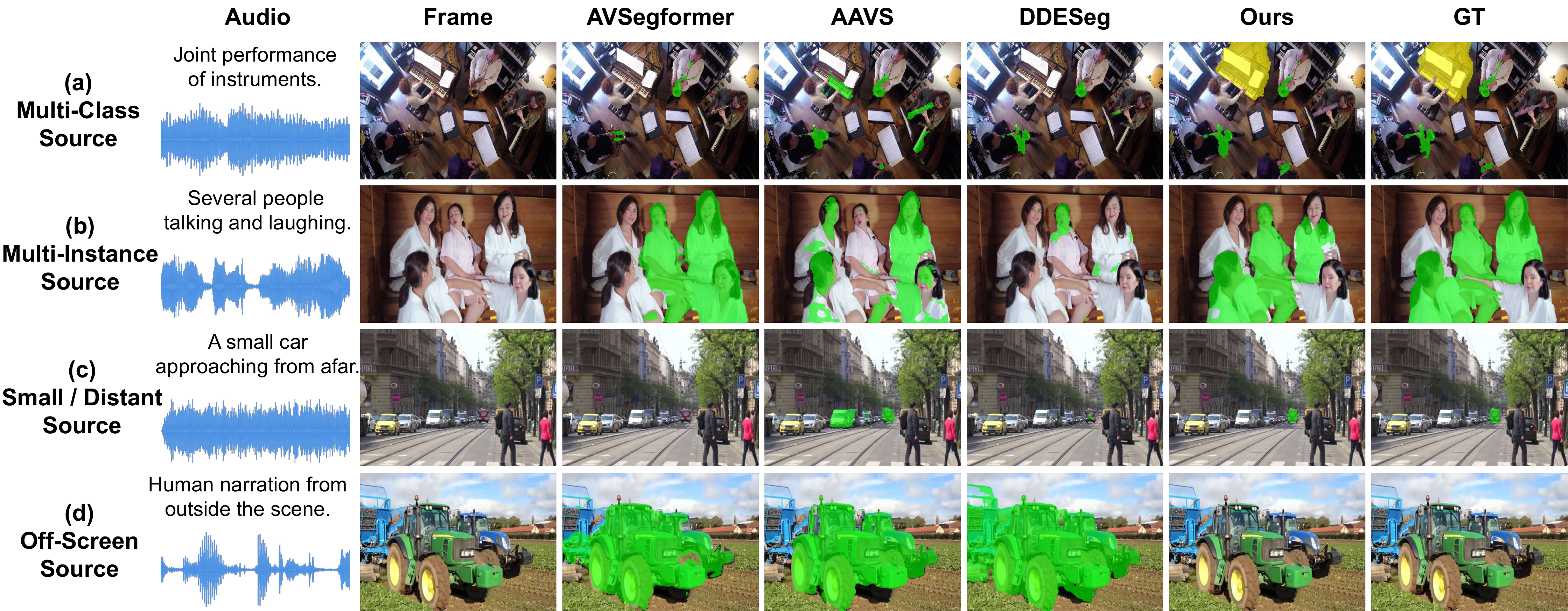}
    \caption{\textbf{Qualitative comparison of DDAVS and previous methods.}
    DDAVS consistently outperforms previous approaches in challenging scenarios involving multiple classes, multiple sources, small or distant sound sources, and off-screen audio cues.}
    \label{teaser}
\end{figure}

Despite its potential, AVS introduces unique challenges illustrated in \cref{teaser}. First, multi-source entanglement in (a) multi-class and (b) multi-instance scenarios prevents precise isolation of individual sound-producers, leading to degraded segmentation performance. Second, audio-visual misalignment hinders cross-modal correspondence; specifically, (c) small or distant sources provide insufficient visual anchors, while (d) off-screen sources lack visual counterparts, often causing spurious activations or incorrect suppression. 

\begin{figure}[t]
    \centering
    \includegraphics[width=1\linewidth]{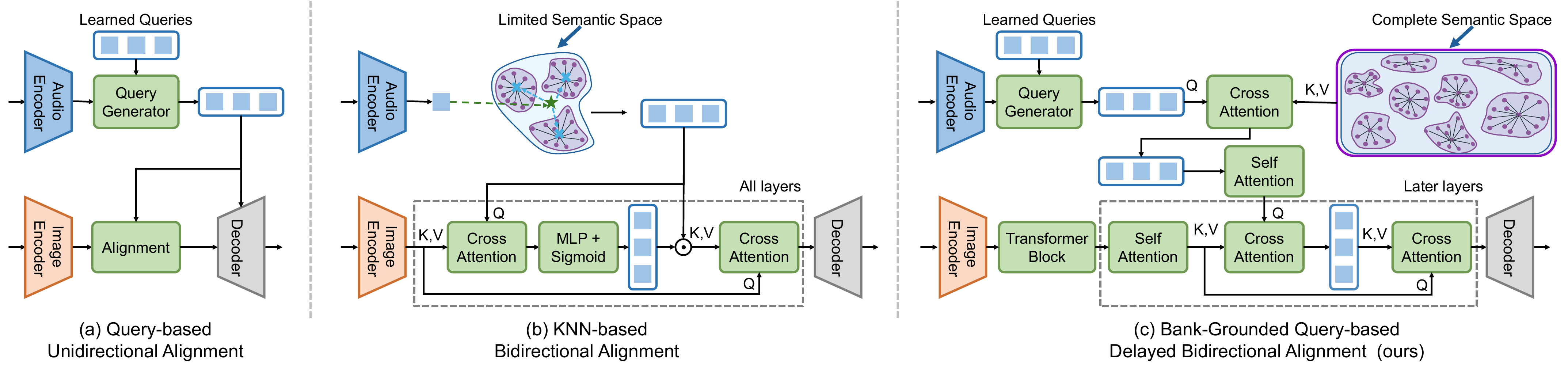}
    \caption{
    \textbf{Comparison of audio-visual segmentation approaches.}
    (a) Query-based audio disentanglement followed by unidirectional audio-to-visual alignment.
    (b) KNN-based audio semantics approximation from a feature bank with gated bidirectional alignment.
    (c) Prototype memory bank-grounded disentanglement with delayed bidirectional cross-attention operating exclusively in later layers.
    }
    \label{fig:method_compare}
\end{figure}

Early approaches resort to an audio disentanglement module using learnable queries to disentangle the audio input into multiple semantics \cite{gao2024avsegformer, li2024qdformer, wang2024avesformer}, followed by unidirectional audio-conditioned visual alignment \cite{wang2024avesformer, zhou2022audio, zhou2024audio} (see \cref{fig:method_compare}(a)). However, this pipeline faces two key limitations: its disentangled semantics reside in a self-organized latent space suboptimal for audio representation, and the unidirectional design prevents visual cues from enhancing scene-aligned audio components or suppressing irrelevant ones (e.g., off-screen sounds), thereby weakening cross-modal refinement.
More recent studies, such as the audio bank-based framework \cite{liu2025dynamic}, attempt to improve semantic clarity by approximating audio semantics via the $K$ nearest centers from a multi-class feature bank (see \cref{fig:method_compare}(b)).
However, when distinct audio sources coexist, weaker-source semantics are often lost due to the constrained semantic space of the $K$ nearest classes, reducing output distinguishability; furthermore, bidirectional alignment relies on a gating mechanism that merely scales audio intensity without aligning to visual semantics or capturing spatial cues. 

In this paper, we present DDAVS, 
a framework for audio-visual segmentation that leverages disentangled audio semantics to enable delayed bidirectional alignment (\cref{fig:method_compare}{(c)}).
Our approach proceeds in two stages. First, in the audio disentanglement stage, we use learnable queries to extract multiple audio semantics and perform cross-attention conditioned on a pre-built multi-class prototype memory bank of single-source audio embeddings.
This anchors the extracted semantics to a structured and stable space, infusing prior knowledge and facilitating subsequent alignment. 
Additionally, we integrate contrastive learning during training to enhance the discriminability and robustness of disentangled audio semantic anchors.
Second, in the alignment stage,  unlike methods that apply cross-attention across all network layers (e.g., \cref{fig:method_compare}(b)), our delayed bidirectional cross-attention operates exclusively in later layers to align audio and visual modalities. The delayed interaction filters low-level noise, while the bidirectional design enables symmetric cross-attention between audio and video, capturing mutual dependencies for precise segmentation.

In summary, our technical contributions are as follows:
\begin{itemize}
\item We propose an AVS framework with delayed bidirectional alignment via disentangled
Audio Semantics for precise segmentation in challenging scenarios such as multi-source, subtle, distant, or off-screen sounds.
\item We propose an audio disentanglement module that anchors query-extracted audio semantics to a prototype memory bank for global consistency, and uses contrastive learning to enhance discriminability and robustness.
\item We propose an audio-visual alignment module using cascaded bidirectional cross-attention to enhance inter-modal interaction and delayed alignment for precise high-level correspondence while reducing low-level noise.
\item Experiments on AVS-Objects and VPO benchmarks demonstrate that DDAVS consistently outperforms prior methods especially in challenging scenarios.
\end{itemize}

\section{Related Work}
\label{sec:related}
\noindent\textbf{Audio-Visual Segmentation.} 
Given an audio signal and an accompanying image or video, audio-visual segmentation aims to produce the segmentation mask of the sounding objects in the image~\cite{liu2023audio, liu2025dynamic, gong2025complementary, ying2025towards, wang2024avesformer, gong2025avs, li2025waveforms, zhou2022audio, zhou2024audio,liu2024open,bai2024self,liu2025stepping}.
As a pioneer, Zhou \etal~\cite{zhou2022audio} propose the audio-visual segmentation problem and introduce the AVSBench benchmark.
Typical AVS methods usually leverage learnable queries~\cite{gao2024avsegformer, li2024qdformer, wang2024avesformer, lv2025consistency, sun2024unveiling, li2023catr, liu2023audio, liu2023audio-visual, ma2024stepping} to extract audio or visual semantics and perform an audio-visual alignment to achieve visual segmentation based on audio cues.
Recent AVS methods focus on text-bridged strategy~\cite{luo2025tavis}, counterfactual learning~\cite{zha2025implicit}, audio enhancement and disentanglement~\cite{liu2025dynamic}, and robust audio-visual alignment~\cite{ huang2025revisiting, mao2025contrastive, liu2025robust}.
In addition to architectural advances, several frameworks use contrastive learning~\cite{chen2020simple, he2020momentum, grill2020bootstrap} to enhance cross-modal alignment and training stability. 
CAVP~\cite{chen2024unraveling}, DiffusionAVS~\cite{mao2025contrastive} and CQFormer~\cite{lv2025consistency} adopt an InfoNCE-based loss~\cite{oord2018representation} to align audio and visual modalities. 
WS-AVS~\cite{mo2023weakly} applies contrastive learning under weak supervision. 
Our method differs from existing approaches by using a bank to anchor and enrich query-generated audio semantics and a delayed bidirectional alignment to guide segmentation. Moreover, we leverage contrastive learning to enhance the discriminability and robustness of audio features rather than to align audio and visual modalities.

\noindent\textbf{Multi-Source Audio Disentanglement.}
In multi-source scenarios, AVS methods usually employ representation-level audio disentanglement mechanisms to separate overlapping sound sources~\cite{liu2025dynamic, li2024qdformer, gao2024avsegformer, wang2024avesformer, ma2024stepping}.
This is often achieved through learnable queries~\cite{li2024qdformer, gao2024avsegformer, wang2024avesformer, ma2024stepping} or $K$-nearest-neighbor-based decomposition~\cite{liu2025dynamic}, where the audio feature is decomposed into multiple audio semantics representing distinct sound emitters. 
However, existing query-based methods produce semantic tokens in a self-organized space without explicit structure.
While the $K$-nearest-neighbor-based method might be limited in discriminability.
We embed audio semantics into an audio-preferred semantic space using a prototype memory bank and enhance their discriminability via contrastive learning.

\noindent\textbf{Audio-Visual Alignment.}
This module establishes spatial and semantic correspondences between the audio and visual modalities before decoding~\cite{wang2024avesformer, zhou2022audio, zhou2024audio, liu2025dynamic, zhou2025aloha, seon2024extending, yang2024cooperation, shi2024cross}. 
Typical AVS methods conduct unidirectional audio-conditioned visual alignment~\cite{wang2024avesformer, zhou2022audio, zhou2024audio}, ignoring the utilization of visual features to improve audio features.
Recent methods~\cite{liu2025dynamic, zhang2025flashvstream, wang2024prompting, seon2024extending, zhang2025thinking,  wang2025ponder, yang2024cooperation, zhang2025alignedgen, gu2025thinking, yang2022lavt} introduce bidirectional alignment to improve inter-modal interaction.
However, the gating mechanism~\cite{liu2025dynamic} and the early alignment~\cite{seon2024extending, wang2024prompting} might hinder effective alignment.
Although AVESFormer~\cite{wang2024avesformer} adopts delayed alignment, the alignment is unidirectional.
In contrast, we propose delayed bidirectional alignment for dynamic cross-modal feature matching, effectively improving segmentation accuracy.

\section{Method}
\label{sec:method}
\begin{figure}[t]
    \centering
    \includegraphics[width=1\linewidth]{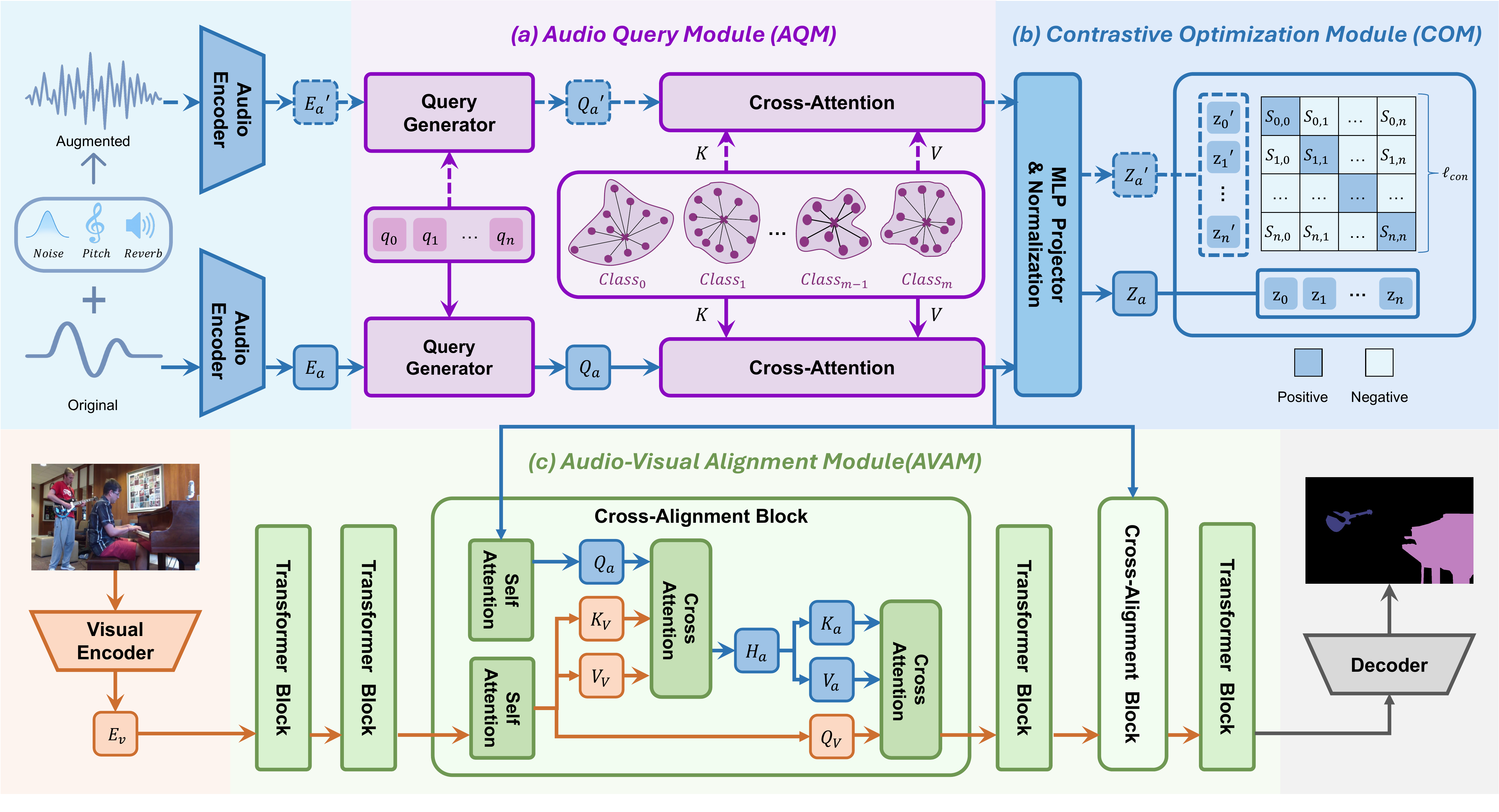}
    \caption{\textbf{Overview of the DDAVS framework.}
    (a) Audio Query Module (AQM) encodes original and augmented waveforms into disentangled semantic queries.
    (b) Contrastive Optimization Module (COM) enhances query robustness through contrastive learning.
    (c) Audio-Visual Alignment Module (AVAM) fuses audio queries with visual features using alignment blocks.
    }
    \label{fig:overview}
\end{figure}

We propose an end-to-end Disentangled Audio Semantics and Delayed Bidirectional Alignment framework (DDAVS). 
As shown in \cref{fig:overview}, the framework comprises three key complementary components: (1) Audio Query Module (AQM) converts audio features into a compact set of disentangled semantic queries anchored to a prototype bank; (2) Contrastive Optimization Module (COM) refines these queries via contrastive learning; and (3) Audio-Visual Alignment Module (AVAM) employs multi-stage dual cross-attention to align both modalities progressively and bidirectionally.
Formally, given a raw audio waveform $A_a$ and its corresponding video clip of frames $I_v$, 
the audio feature $E_a = \mathcal{E}_a(A_a)$ and visual feature $E_v = \mathcal{E}_v(I_v)$ are extracted by their encoders $\mathcal{E}_a$ and $\mathcal{E}_v$. 
$H_v^i$ denotes the visual feature after the $i$-th decoder stage.
The inference pipeline is:
\begin{align}
Q &=\text{AQM}(E_a) \\
H_v^0 &= E_v \\
H_v^i &= \text{AVAM}^i(Q, H_v^{i-1}), \quad i=1,\dots,L
\label{eq:workflow}
\end{align}
The AQM transform $E_a$ into a disentangled representation $Q$, while $E_v$ serves as the initial visual input $H_v^0$. 
The COM is only used during training to provide an additional contrastive loss.
DDAVS then performs $L$ iterative stages of alignment and fusion through the Audio-Visual Alignment Module (AVAM), each applying dual cross-attention followed by Transformer refinement 
to synchronize and integrate the two modalities.
This progressive alignment yields increasingly discriminative and spatially coherent representations.
Finally, a lightweight decoder $\mathcal{D}$ generates the pixel-level segmentation mask $\hat{Y} = \mathcal{D}(H_v^{L})$ that highlights audible regions within the scene.

\subsection{Audio Query Module}
\label{subsec:aqm}

The Audio Query Module (AQM) transforms the encoded audio features $E_a \in \mathbb{R}^{L\times d}$ into compact and disentangled representations by learned queries. 
It aims to decouple overlapping sound sources and map them into a stable semantic space anchored by a global prototype memory bank.

\noindent \textbf{Query Generation.}
As shown in \cref{fig:overview}, the \textit{Query Generator} is implemented 
with a Q-Former~\cite{li2023blip2}, 
which maps the sequential audio tokens $E_a$ into $n$ audio query vectors: $Q_a = f_{\text{QG}}(E_a; \{q_i\}_{i=1}^n) \in \mathbb{R}^{n\times d}$.
Each learnable query $q_i$ is a latent slot that focuses on a distinct sound component, allowing AQM to separate co-occurring acoustic patterns.

\noindent \textbf{Bank Construction.}
We construct a global prototype memory bank $\mathcal{M}$ to provide stable semantic anchors for query refinement. For each class $i \in \{1,\dots,C\}$, we collect single-source audio clips where class $i$ is the only audible sound, extract features using the HTSAT backbone, and obtain embeddings $E_i = \{x_k^i \in \mathbb{R}^d\}_{k=1}^{n_i}$. Applying K-means++ clustering with $K_i$ clusters to $E_i$, we select the cluster centroids as class-specific prototypes $C_i = \{c_{i,j} \in \mathbb{R}^d\}_{j=1}^{K_i}$ . Concatenating all class prototypes yields the global bank $\mathcal{M} = \text{concat}(C_1,\dots,C_C) \in \mathbb{R}^{K_i \times d}$. Crucially, $\mathcal{M}$ remains fixed during both training and inference, ensuring consistent semantic grounding across all samples. Further implementation details are provided in the supplementary material.

\noindent \textbf{Bank-Guided Refinement.}
The initial audio queries $Q_a$ are refined through cross-attention with the prototype memory bank $\mathcal{M}$ described above. Specifically, $Q_a$ interacts with $\mathcal{M}$ via:
% \begin{equation}
\begin{align}
% \begin{equation}
A &= \operatorname{Softmax}\left( \frac{ (Q_a W_Q) (\mathcal{M} W_K)^{\top} }{ \sqrt{d} } \right) \\
\widetilde{Q} &= A (\mathcal{M} W_V) \\
Q &= \operatorname{LN}( Q_a + \gamma \widetilde{Q} )
\end{align}
where $W_{Q/K/V} \in \mathbb{R}^{d\times d}$ are projection layers, $\gamma$ is a scaling factor, and $\operatorname{LN}$ denotes layer normalization. This process anchors each query to its most relevant semantic prototype, injecting class-aware prior knowledge while preserving query diversity. $Q$ denotes the final bank-grounded audio queries used for alignment.

While AQM effectively generates semantically aligned queries, their embeddings remain insufficiently discriminative and tend to be dominated by salient audio sources. In multi-sound scenarios (e.g., speech co-occurring with engine noise), dominant components often suppress weaker signals, resulting in poor inter-class separation. This limitation motivates the Contrastive Optimization Module (COM) introduced next.

\subsection{Contrastive Optimization Module}
\label{subsec:com}
To mitigate the issue of insufficiently discriminative audio embeddings, we design a Contrastive Optimization Module (COM), which employs contrastive learning to enhance semantic separation between different sound classes and improve robustness to acoustic variations.

\noindent \textbf{Audio Signal Augmentation.}
To improve robustness under acoustic perturbations, we apply waveform-level augmentations to the raw audio signal. The pipeline first resamples audio to $16$\,kHz, center-crops or pads to a fixed duration, and normalizes to $[-1,1]$. We then generate an augmented counterpart $A_a' = g(A_a)$ using WavAugment \cite{kharitonov2021data,jiang2020speech}, where $g$ applies a chain of time-domain effects: reverberation ($r \in [20,40]$), pitch shift ($\Delta p \in [-150,150]$ cents), dynamic-range compression, and volume jitter ($\mathrm{SNR} \in [10,20]$ dB). 
% Parameter ranges are summarized in \cref{tab:aug_cfg}. 
Parameter ranges are summarized in supplementary. 
Crucially, only COM processes the augmented branch: the clean waveform $A_a$ drives the main segmentation path via $E_a = \mathcal{E}_a(A_a)$, while the augmented waveform $A_a'$ is exclusively used by COM to produce $E_a' = \mathcal{E}_a(A_a')$ and corresponding query set $Q'$. This design ensures semantic content remains intact while introducing moderate acoustic variations for robust representation learning.

\noindent \textbf{Contrastive Learning.}
Given the refined query set $Q=\{q_i\}_{i=1}^{n}$ and its augmented counterpart $Q'=\{q_i'\}_{i=1}^{n}$,
we apply a projector $\phi(\cdot)$ and $\ell_2$-normalization to each query:
\begin{equation}
    z_i = \frac{\phi(q_i)}{\|\phi(q_i)\|_2}, \qquad
    z_i' = \frac{\phi(q_i')}{\|\phi(q_i')\|_2}
\end{equation}
Let $s_{i,j} = z_i^\top z_j'$.
The contrastive loss is defined as:
\begin{equation}
    \mathcal{L}_{\text{con}}
     = -\frac{1}{n}\sum_{i=1}^{n}
       \log
       \frac{\exp(s_{i,i}/\tau)}
        {\sum_{j=1}^{n}\exp(s_{i,j}/\tau)}
    \label{eq:lcon}
\end{equation}
where $\tau$ is the temperature coefficient.
$\mathcal{L}_{\text{con}}$ pulls together positive pairs $(z_i,z_i')$ and pushes apart negative pairs $\{(z_i,z_j')\}_{j\neq i}$, enlarging inter-query margins under acoustic variations.
After contrastive optimization, we obtain enhanced audio embeddings $Q=\{q_i\}_{i=1}^{n}$that are more robust to noise and better disentangled across sound classes, directly strengthening downstream cross-modal alignment and segmentation stability.

\subsection{Audio-Visual Alignment Module}
\label{subsec:vafm}

The Audio-Visual Alignment Module (AVAM) aligns visual and auditory modalities to precisely localize sound-producing regions. As illustrated in \cref{fig:overview}, AVAM employs an alternating architecture of Cross Alignment Blocks and Transformer Blocks, progressively refining spatial coherence and cross-modal interactions.

Within the i-th Cross Alignment Block, given  
hidden state $H_v^{i-1}$
and enhanced audio queries $Q$, alignment begins with audio queries attending to visual tokens. This design leverages the complementary nature of the modalities: audio delivers concise semantic cues about \textit{what} is sounding, while vision supplies the spatial context for \textit{where} it originates. Consequently, this directional attention naturally guides the model toward sound-relevant regions and suppresses background distractions without explicit supervision.
\begin{align}
    \tilde{H}_a^{i} &= \mathrm{SelfAttn}(Q) 
    \label{eq:audio_sa} \\
    \tilde{H}_v^{i} &= \mathrm{SelfAttn}(H_v^{i-1}) 
    \label{eq:visual_sa} \\
    H_a^{i} &= \mathrm{CrossAttn}(\tilde{H}_a^{i},\, \tilde{H}_v^{i},\, \tilde{H}_v^{i}) 
    \label{eq:audio_xa} \\
    H_v^{i} &= \mathrm{CrossAttn}(\tilde{H}_v^{i},\, H_a^{i},\, H_a^{i})
    \label{eq:visual_xa}
\end{align}
\noindent \textbf{Audio-Guided Filtering.}
Audio queries attend to visual tokens to extract sound-relevant visual evidence as shown in \Cref{eq:audio_sa,eq:audio_xa}.
Here $\mathrm{SelfAttn}(x)$ and $\mathrm{CrossAttn}(x_Q, x_K, x_V)$
denote standard transformer self-attention and cross-attention.
This step generates audio-conditioned visual features focused on regions visually correlated with emitted sounds.
Using visual features from pretrained encoder as keys and values simultaneously constrains the relatively noisier audio representations, providing an implicit denoising effect.

\noindent \textbf{Visual-Guided Enhancement.}
The updated audio representations then act as keys and values to inject discriminative acoustic cues back into the visual stream following \Cref{eq:visual_sa,eq:visual_xa}.
This reverse attention leverages the purified acoustic cues to further sharpen visual activations at sound-producing locations, completing a robust bidirectional alignment cycle. Ablation on this bidirectional ordering is provided in the supplementary material.

\begin{figure}[t]
    \centering    
    \includegraphics[width=0.8\linewidth]{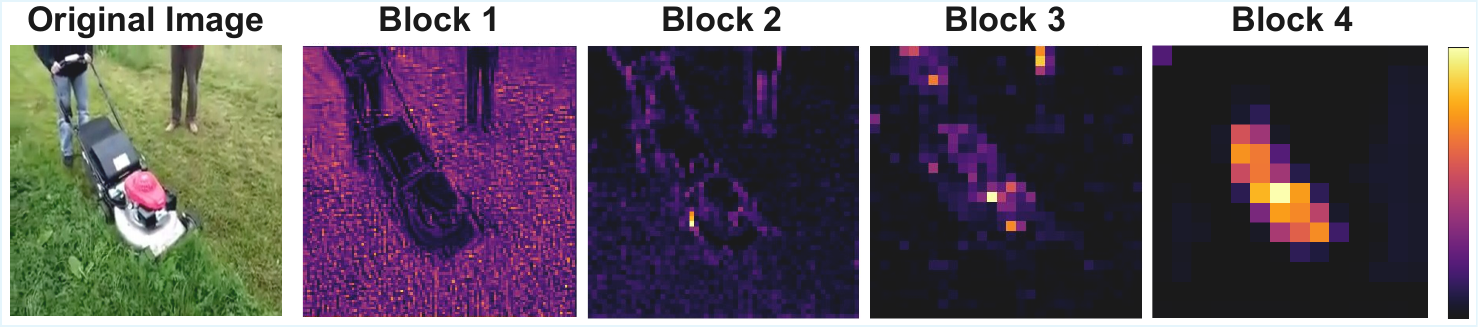}
    \caption{\textbf{Attention maps of audio-injected Transformer blocks across layers.}
    It is observed that injecting audio features into block 3 and 4 bringing clearer instance-level attention, compared to the blurry pattern at earlier blocks.
}
    \label{fig:attn_map}
\end{figure}

\noindent \textbf{Delayed Cross-Modal Alignment.}
Cross-modal Alignment is applied exclusively between the third and fourth layers of the four-block architecture. As shown in \cref{fig:attn_map}, early fusion captures fragmented pixel patterns, while deeper fusion highlights coherent region- and instance-level structures essential for audio-visual grounding. Quantitative results in \cref{tab:audio_inject_blocks} confirm the block-3-and-4 configuration achieves peak $\mathcal{J}\&\mathcal{F}$ performance. 
The resulting audio-conditioned visual features $H_v^L$ are then decoded into the segmentation mask $\hat{Y} = \mathcal{D}(H_v^L)$.

\subsection{Optimization}
\label{sec:3.4}
We train the DDAVS model with a unified objective:
\begin{equation}
    \begin{aligned}
        \mathcal{L}_{\mathrm{total}}
        &= \lambda_{\mathrm{ce}}\mathcal{L}_{\mathrm{ce}}
         + \lambda_{\mathrm{dice}}\mathcal{L}_{\mathrm{dice}} 
         + \lambda_{\mathrm{iou}}\mathcal{L}_{\mathrm{iou}}
         + \lambda_{\mathrm{con}}\mathcal{L}_{\mathrm{con}}.
    \end{aligned}
    \label{eq:total}
\end{equation}
The cross-entropy loss $\mathcal{L}_{\mathrm{ce}}$ provides pixel-wise supervision, while the Dice and IoU losses
$\mathcal{L}_{\mathrm{dice}}$ and $\mathcal{L}_{\mathrm{iou}}$ encourage region completeness and accurate boundary alignment.
Beyond these segmentation losses, the contrastive term $\mathcal{L}_{\mathrm{con}}$ (See \cref{eq:lcon})
enforces discriminative audio queries by enlarging inter-query margins under acoustic perturbations.

The segmentation losses (Cross-entropy, Focal, Dice, IoU) and  $\lambda$ coefficients are detailed in the supplementary material.

\begin{table}[!t]
    \caption{\textbf{Quantitative comparisons on the AVSBench dataset}, including single-source (AVS-Objects-S4), multi-source (AVS-Objects-MS3), and semantic-source (AVS-Semantic) settings.
    Best results in \textbf{bold}, while second best \underline{underlined}.}
    \label{tab:avs_results}
    \scriptsize
    \centering
    \setlength{\tabcolsep}{5pt}
    \begin{tabular}{l|ccc|ccc|ccc}
        \toprule
        \textbf{Method} 
        & \multicolumn{3}{c|}{\textbf{AVS-Objects-S4 }} 
        & \multicolumn{3}{c|}{\textbf{AVS-Objects-MS3}} 
        & \multicolumn{3}{c}{\textbf{AVS-Semantic}} \\ 
        \cmidrule{2-10}
        & $\mathcal{J}\&\mathcal{F}$ & $\mathcal{J}$ & $\mathcal{F}$
        & $\mathcal{J}\&\mathcal{F}$ & $\mathcal{J}$ & $\mathcal{F}$
        & $\mathcal{J}\&\mathcal{F}$ & $\mathcal{J}$ & $\mathcal{F}$ \\ 
        \midrule
        TPAVI~\cite{zhou2022audio}~\textcolor{gray}{\tiny{[ECCV22]}}
        & 83.3 & 78.7 & 87.9
        & 59.3 & 54.0 & 64.5
        & 32.5 & 29.8 & 35.2 \\
        
        CATR~\cite{li2023catr}~\textcolor{gray}{\tiny{[ACM-MM23]}}
        & 87.9 & 84.4 & 91.3
        & 68.6 & 62.7 & 74.5
        & 35.7 & 32.8 & 38.5 \\
        
        AuTR~\cite{liu2023audio}~\textcolor{gray}{\tiny{[Arxiv23]}}
        & 82.1 & 77.6 & 86.5
        & 72.0 & 66.2 & 77.7
        & -- & -- & -- \\
        
        AVSC~\cite{liu2023audio-visual}~\textcolor{gray}{\tiny{[ACM-MM23]}}
        & 85.0 & 81.3 & 88.6
        & 62.6 & 59.5 & 65.8
        & -- & -- & -- \\
        
        ECMVAE~\cite{mao2023multimodal}~\textcolor{gray}{\tiny{[ICCV23]}}
        & 85.9 & 81.7 & 90.1
        & 64.3 & 57.8 & 70.8
        & -- & -- & -- \\
        
        AQFormer~\cite{huang2023discovering}~\textcolor{gray}{\tiny{[IJCAI23]}}
        & 85.5 & 81.6 & 89.4
        & 67.5 & 62.2 & 72.7
        & -- & -- & -- \\
        
        BAVS~\cite{liu2024bavs}~\textcolor{gray}{\tiny{[TMM24]}}
        & 86.2 & 82.7 & 89.8
        & 62.8 & 59.6 & 65.9
        & 35.6 & 33.6 & 37.5 \\
        
        AVSegFormer~\cite{gao2024avsegformer}~\textcolor{gray}{\tiny{[AAAI24]}}
        & 86.8 & 83.1 & 90.5
        & 67.2 & 61.3 & 73.0
        & 40.1 & 37.3 & 42.8 \\
        
        GAVS~\cite{wang2024prompting}~\textcolor{gray}{\tiny{[AAAI24]}}
        & 85.1 & 80.1 & 90.0
        & 70.6 & 63.7 & 77.4
        & -- & -- & -- \\
        
        AVSBG~\cite{hao2024improving}~\textcolor{gray}{\tiny{[AAAI24]}}
        & 86.1 & 81.7 & 90.4
        & 61.0 & 55.1 & 66.8
        & -- & -- & -- \\
        
        AVESFormer~\cite{wang2024avesformer}~\textcolor{gray}{\tiny{[Arxiv24]}}
        & 84.5 & 79.9 & 89.1
        & 63.3 & 57.9 & 68.7
        & 34.0 & 31.2 & 36.8 \\
        
        QDFormer~\cite{li2024qdformer}~\textcolor{gray}{\tiny{[CVPR24]}}
        & 83.9 & 79.5 & 88.2
        & 64.0 & 61.9 & 66.1
        & -- & -- & -- \\
        
        CAVP~\cite{chen2024unraveling}~\textcolor{gray}{\tiny{[CVPR24]}}
        & 83.8 & 78.8 & 88.9
        & 61.5 & 55.8 & 67.1
        & 32.8 & 30.4 & 35.3 \\
        
        COMBO~\cite{Yang_2024_CVPR}~\textcolor{gray}{\tiny{[CVPR24]}}
        & 88.3 & 84.7 & 91.9
        & 65.2 & 59.2 & 71.2
        & 44.1 & 42.1 & 46.1 \\
        
        AAVS~\cite{ma2024stepping}~\textcolor{gray}{\tiny{[ECCV24]}}
        & 87.3 & 83.2 & 91.3
        & 72.5 & 67.3 & 77.6
        & \underline{50.9} & \underline{48.5} & \underline{53.2} \\
        
        CPM~\cite{chen2024cpm}~\textcolor{gray}{\tiny{[ECCV24]}}
        & 85.9 & 81.4 & 90.5
        & 65.4 & 59.8 & 71.0
        & 37.1 & 34.5 & 39.6 \\
        
        BiasAVS~\cite{sun2024unveiling}~\textcolor{gray}{\tiny{[ACM-MM24]}}
        & 88.2 & 83.3 & 93.0
        & 74.0 & 67.2 & \underline{80.8}
        & 47.2 & 44.4 & 49.9 \\
        DiffusionAVS~\cite{mao2025contrastive}~\textcolor{gray}{\tiny{[TIP25]}}
        &85.9 & 81.5 & 90.3
        &65.4 & 59.6 & 71.2
        &40.6 & 38.1 & 43.0 \\

        VCT~\cite{huang2025revisiting}~\textcolor{gray}{\tiny{[CVPR25]}}
        & 88.5 & 84.7 & 92.3
        & 73.4 & 67.5 & 79.3
        & 50.4 & 47.9 & 52.9 \\
        
        DDESeg~\cite{liu2025dynamic}~\textcolor{gray}{\tiny{[CVPR25]}}
        & \underline{91.1} & \underline{89.1} & 93.1
        & 72.2 & 68.1 & 76.2
        & 49.6 & 47.1 & 52.1 \\
        
        TAViS~\cite{luo2025tavis}~\textcolor{gray}{\tiny{[ICCV25]}}
        & 88.0 & 84.8 & 91.2
        & 72.1 & 68.2 & 75.9
        & -- & 44.2 & -- \\
        
        ICF~\cite{zha2025implicit}~\textcolor{gray}{\tiny{[ICCV25]}}
        & 90.1 & 86.6 & \underline{93.5}
        & 69.9 & 64.4 & 75.4
        & 48.2 & 45.0 & 51.3 \\
    
        CCFormer~\cite{gong2025complementary}~\textcolor{gray}{\tiny{[TMM25]}}
        & 88.8 & 84.9 & 92.7
        & \underline{75.6} & \underline{70.7} & 80.5
        & 45.4 & 42.3 & 48.5 \\
        
        \midrule
        \textbf{DDAVS (Ours)} 
        & \textbf{92.4} 
        & \textbf{90.6} 
        & \textbf{94.2}
        & \textbf{76.0}
        & \textbf{70.9}
        & \textbf{81.1}
        & \textbf{52.9}
        & \textbf{50.2}
        & \textbf{55.6} \\
        \bottomrule
    \end{tabular}
\end{table}

\section{Experiments}

\noindent\textbf{Implementation Details.}
The visual backbone is initialized from MiT-B5~\cite{xie2021segformer}, and the audio encoder adopts HTSAT~\cite{chen2022hts} pretrained on AudioSet~\cite{gemmeke2017audio}. 
Following DDESeg~\cite{liu2025dynamic}, we construct the prototype memory bank from single-sounding source signals. Specifically, we build the bank from training splits by filtering 12356/12202 single-source clips across 71/21 categories for AVS-Semantic/VPO-SS, respectively.
For each category, HTSAT embeddings are clustered via K-means++, and the $K_c=5$ features nearest to the cluster center are fixed as prototypes, yielding 355/105 anchors in total.
All experiments are conducted on a workstation equipped with eight NVIDIA RTX 4090 GPUs (24 GB each). Training uses the AdamW optimizer with an initial learning rate of $1\times10^{-4}$ and a batch size of 64. We also fix random seeds to ensure reproducibility.

\noindent\textbf{Datasets and Metrics.}
We evaluate DDAVS on two audio-visual segmentation benchmarks: AVSBench~\cite{zhou2022audio, zhou2024audio} and VPO~\cite{chen2024unraveling}, which cover single-source, multi-source, and semantic conditions.
Following common practice~\cite{zhou2022audio, chen2024unraveling} in AVS, 
we adopt the Jaccard index ($\mathcal{J}$), the F-score ($\mathcal{F}$) and their average $\mathcal{J}\&\mathcal{F}$ as evaluation metrics. 
The F-score is
$\mathcal{F} = \frac{(1+\beta^{2}) \cdot \text{Precision} \cdot \text{Recall}}{\beta^{2} \cdot \text{Precision} + \text{Recall}},
$ 
where $\beta^{2}=0.3$, which places more emphasis on recall. 
For AVSBench (including AVS-Object and AVS-Semantic), the scores are computed using the official TPAVI evaluation protocol~\cite{zhou2022audio}, 
while for VPO we follow the metric implementation of CAVP~\cite{chen2024unraveling}.

\subsection{Quantitative Evaluation}
\noindent \textbf{AVSBench.}
\cref{tab:avs_results} presents the experimental results on AVSBench.
DDAVS achieves state-of-the-art performance across all subsets.
On the semantic subset AVSS involving spatial and categorical ambiguity, DDAVS improves over previous best baseline by 2.0\% $\mathcal{J}\&\mathcal{F}$, indicating that disentangled audio queries and dual-stage fusion effectively reduce interference between overlapping sources.

\noindent \textbf{VPO.}
\cref{tab:vpo_results} presents the experimental results on VPO.
DDAVS outperforms the state-of-the-art by 1.81\% and 3.54\% $\mathcal{J}\&\mathcal{F}$  on VPO-MS (multi-source) and VPO-MSMI (multi-source and multi-instance).
These results demonstrate that the advantage of DDAVS in complex multi-source scenes that require robust cross-modal representation, while it also remains competitive under simple conditions such as VPO-SS.

\begin{table}[t]
    \caption{\textbf{Quantitative comparisons on the VPO dataset}, including single-source (VPO-SS), multi-source (VPO-MS), and multi-source multi-instance (VPO-MSMI) settings. 
    Best results in \textbf{bold}, while second best \underline{underlined}.}
    \vspace{-4pt}
    \label{tab:vpo_results}
    \centering
    \scriptsize
    \setlength{\tabcolsep}{2.6pt}
    \begin{tabular}{l|ccc|ccc|ccc}
    \toprule
    \textbf{Method} 
    & \multicolumn{3}{c|}{\textbf{VPO-SS}} 
    & \multicolumn{3}{c|}{\textbf{VPO-MS}} 
    & \multicolumn{3}{c}{\textbf{VPO-MSMI}} \\ 
    \cmidrule{2-10}
    & $\mathcal{J}\&\mathcal{F} \uparrow$ & $\mathcal{J} \uparrow$ & $\mathcal{F} \uparrow$
    & $\mathcal{J}\&\mathcal{F} \uparrow$ & $\mathcal{J} \uparrow$ & $\mathcal{F} \uparrow$
    & $\mathcal{J}\&\mathcal{F} \uparrow$ & $\mathcal{J} \uparrow$ & $\mathcal{F} \uparrow$ \\
    \midrule
    
    TPAVI~\cite{zhou2022audio}~\textcolor{gray}{\tiny{[ECCV22]}}
    & 44.63 & 41.64 & 47.62 
    & 45.68 & 42.30 & 49.06 
    & 43.19 & 40.03 & 46.34 \\
    
    AVSegFormer~\cite{gao2024avsegformer}~\textcolor{gray}{\tiny{[AAAI24]}}
    & 45.94 & 43.81 & 48.06 
    & 43.72 & 47.30 & 40.14 
    & 49.93 & 47.19 & 52.67 \\
    
    CAVP~\cite{chen2024unraveling}~\textcolor{gray}{\tiny{[CVPR24]}}
    & 67.02 & 58.81 & 75.23 
    & 61.32 & 53.24 & 69.39 
    & 56.48 & 48.18 & 64.78 \\
    
    BiasAVS~\cite{sun2024unveiling}~\textcolor{gray}{\tiny{[ECCV24]}}
    & 67.46 & 59.14 & 75.78 
    & 63.42 & 55.61 & 71.23 
    & 57.94 & 49.60 & 66.27 \\
    
    CPM~\cite{chen2024cpm}~\textcolor{gray}{\tiny{[ECCV24]}}
    &73.49 & 67.09 & 79.88
    &72.91 & 65.91 & 79.90 
    &68.07 & 60.55 & 75.58 \\
    
    AAVS~\cite{ma2024stepping}~\textcolor{gray}{\tiny{[ACM-MM24]}}
    & 68.54 & 59.72 & 77.35 
    & 64.26 & 56.23 & 72.29 
    & 58.76 & 50.11 & 67.40 \\
    
    RAVS~\cite{liu2025robust}~\textcolor{gray}{\tiny{[CVPR25]}}
    & \underline{74.97} & \underline{68.03} & \underline{81.90} 
    & 73.49 & 66.97 & 80.01 
    & \underline{69.30} & 61.89 & \underline{76.70} \\
    
    DDESeg~\cite{liu2025dynamic}~\textcolor{gray}{\tiny{[CVPR25]}}
    & 74.38 & 67.55 & 81.20 
    & \underline{74.30} & \underline{67.64} & \underline{80.96} 
    & 68.39 & \underline{62.11} & 74.67 \\
    \midrule
    \textbf{DDAVS} 
    & \textbf{75.03}
    & \textbf{68.06}
    & \textbf{81.99}
    & \textbf{76.11}
    & \textbf{69.61}
    & \textbf{82.60}
    & \textbf{72.84}
    & \textbf{65.96}
    & \textbf{79.72} \\
    \bottomrule
    \end{tabular}
    \vspace{-2pt}
\end{table}

\begin{figure}[!h]
    \centering
    \includegraphics[width=1\linewidth]{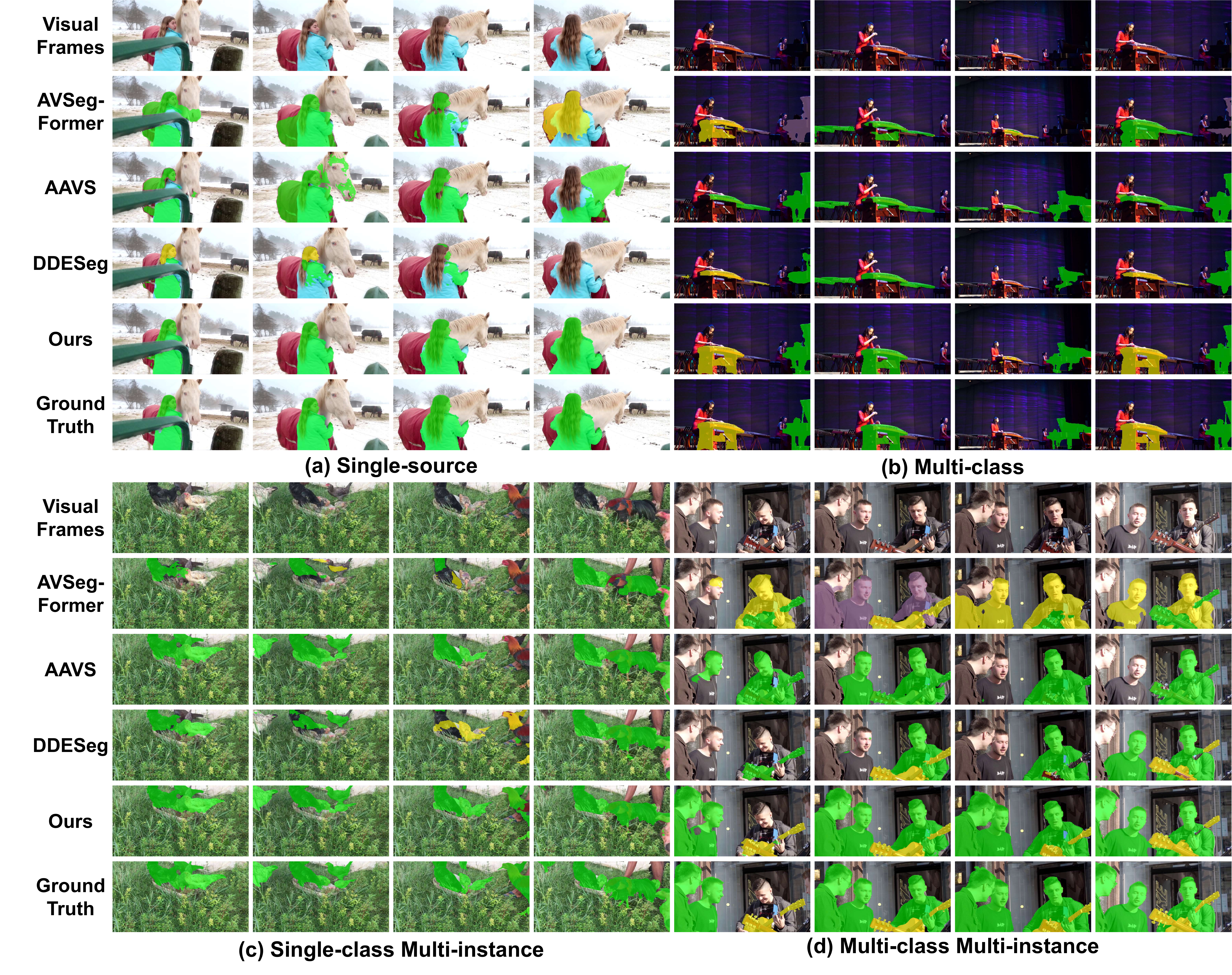}
    \caption{\textbf{Qualitative results on AVSBench.}
        DDAVS produces cleaner and more source-consistent masks than previous baselines AVSegFormer, AAVS, and DDESeg, especially in complex multi-source scenes (multi-class, multi-instance).
    }
    \label{fig:qua_avss}
\end{figure}

\begin{figure}[t]
    \centering
    \includegraphics[width=1\linewidth]{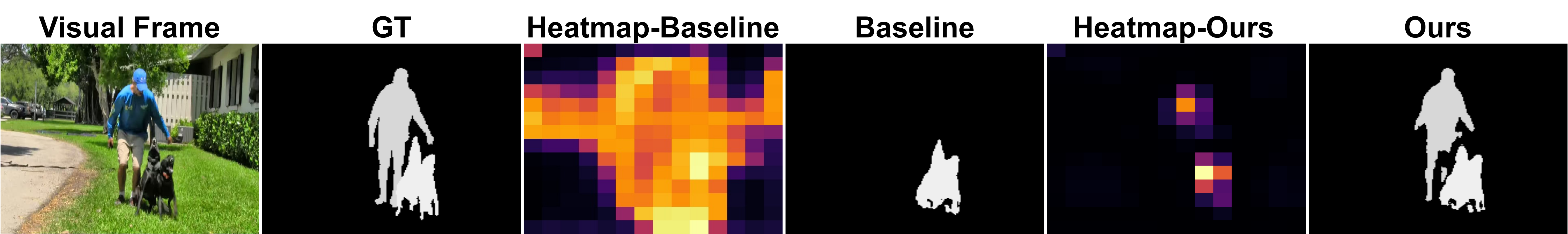}
    \caption{\textbf{Weak target localization.} 
    DDAVS precisely activates both dominant (dog) and weak (person) targets, while the baseline is biased toward the louder sound.
    }
    \label{fig:weak_sound}
\end{figure}

\begin{table}[t]
    \centering
    % \scriptsize
    \setlength{\tabcolsep}{3pt} 
    \caption{\textbf{Quantitative component ablation.} Performance and efficiency metrics of DDAVS variants on AVS-MS3 and AVSS.}
    \vspace{-4pt}
    \label{tab:combined_ablation}
    \begin{tabular}{l | c | c | ccc | ccc}
        \toprule
        \multirow{2}{*}{Method} & \multirow{2}{*}{Params} & \multirow{2}{*}{FLOPs} & \multicolumn{3}{c|}{\textit{AVS-MS3}} & \multicolumn{3}{c}{\textit{AVSS}} \\
        \cmidrule{4-9} 
        & & & $\mathcal{J}\&\mathcal{F}$ & $\mathcal{J}$ & $\mathcal{F}$ & $\mathcal{J}\&\mathcal{F}$ & $\mathcal{J}$ & $\mathcal{F}$ \\
        \midrule
        Baseline & 129.02M & 83.56G & 69.71 & 65.88 & 73.54 & 48.63 & 45.83 & 51.42 \\
        +AQM & 152.06M & 85.39G & 71.89 & 68.04 & 75.73 & 49.80 & 46.73 & 52.86 \\
        +AQM+COM & 152.06M & 85.39G & 74.07 & 69.32 & 78.81 & 51.70 & 48.56 & 54.83 \\
        +AVAM & 127.24M & 83.91G & 73.16 & 68.96 & 77.36 & 51.45 & 48.13 & 54.76 \\
        +AQM+AVAM & 150.29M & 85.72G & 73.75 & 69.06 & 78.44 & 51.52 & 48.25 & 54.79 \\
        \textbf{DDAVS} & 150.29M & 85.72G & \textbf{76.01} & \textbf{70.92} & \textbf{81.10} & \textbf{52.94} & \textbf{50.20} & \textbf{55.67} \\
        \bottomrule
    \end{tabular}
    \vspace{-10pt}
\end{table} 

\begin{figure}[!h]
    \centering
    \includegraphics[width=1\linewidth]{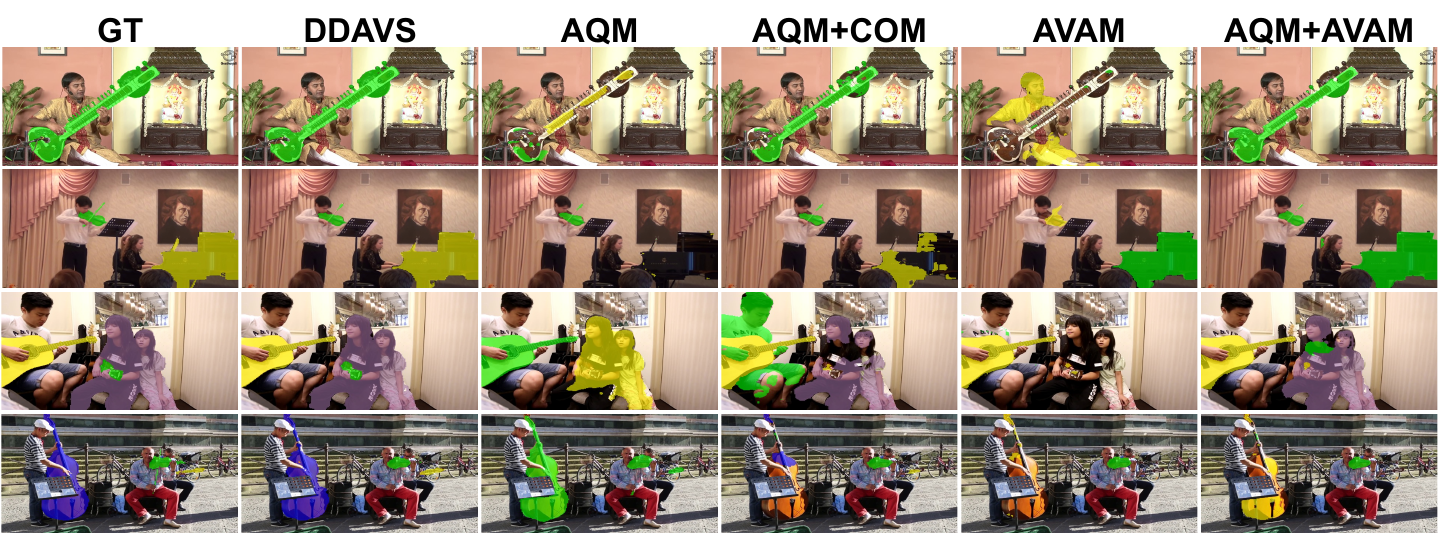}
    \caption{\textbf{Qualitative component ablation.} Visual comparison of segmentation masks: AQM improves localization, COM improves audio-query robustness, AVAM strengthens spatial alignment. Full model achieves precise multi-source segmentation.}
    \label{fig:component-ablation}
\end{figure}

\subsection{Qualitative Evaluation}
\label{sec:qua_avss}
\noindent\textbf{Qualitative Results.}
As shown in \cref{fig:qua_avss}, DDAVS produces cleaner and more precise segmentation masks than baselines across diverse scenarios:
% Across diverse scenes, DDAVS consistently produces cleaner and more source-consistent masks than previous methods. 
(a) isolates the speaking human and suppresses the silent horse while prior methods leak activation to the horse.  
(b) segments all sounding instruments(e.g., saxophone, guitar) without activating silent objects whereas others miss sources or misassign segments.  
(c) distinctly identifies each person and their guitar while competing models merge individuals or omit instruments.  
(d) robustly handles complex multi-person multi-instrument scenes under occlusion while baselines exhibit misassignment and fragmentation.  
consistently demonstrates superior multi-source disentanglement and mask fidelity.

\noindent\textbf{Robustness in Multi-Source Scenarios.}
We evaluate weak target localization under acoustic imbalance. As shown in \cref{fig:weak_sound}, the baseline focuses diffusely on the dominant sound (dog) and misses the weak target (person), whereas DDAVS generates sharp, complete activations for both sources, confirming its robust disentanglement of overlapping audio-visual signals.

\begin{figure}[t]
    \centering
    \begin{minipage}[b]{0.45\linewidth}
        \centering
        \makeatletter\def\@captype{figure}\makeatother 
        \includegraphics[width=\linewidth]{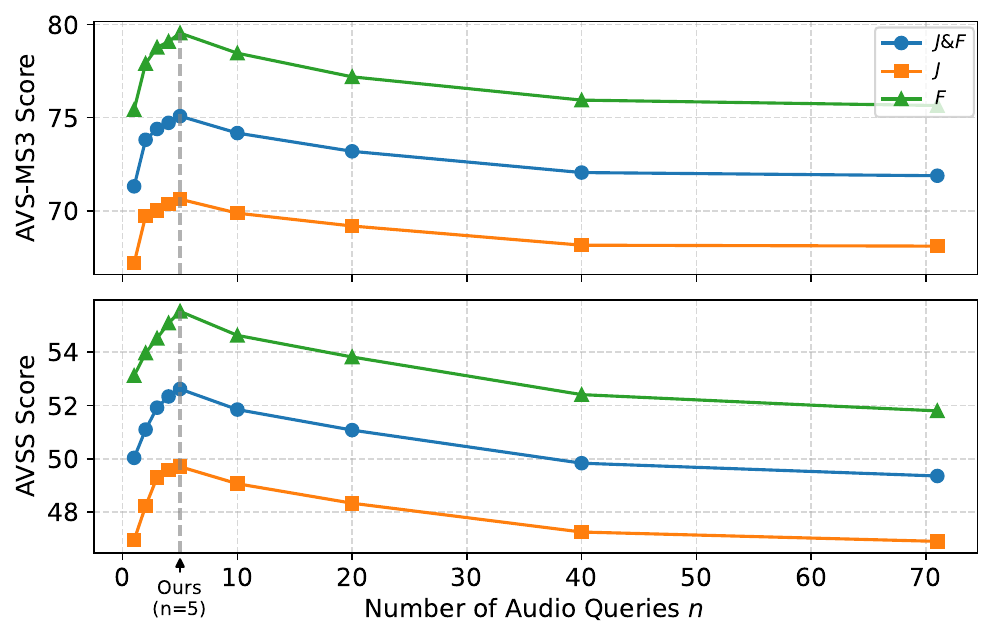}
        \vspace{-20pt}
        \caption{\textbf{Effect of audio query number.} 
        $n=5$ yields the best results, while higher values cause degradation.}
        \vspace{-45pt}
        \label{fig:query_ablation}
    \end{minipage}
    \hfill 
    \begin{minipage}[b]{0.51\linewidth}
        \centering
        \makeatletter\def\@captype{table}\makeatother 
        \footnotesize
         \setlength{\tabcolsep}{2pt}
        \caption{\textbf{Out-of-bank robustness}. Performance when removing $x\%$ of class prototypes in the memory bank.}
        \label{tab:oob}
        \begin{tabular}{lcc}
            \toprule
            Setting & \textit{MS3} ($\mathcal{J}\&\mathcal{F}$) & \textit{AVSS} ($\mathcal{J}\&\mathcal{F}$) \\
            \midrule
            Baseline       & 69.71 & 48.63 \\
            OOB-100\%      & 73.16 & 51.45 \\
            OOB-80\%       & 73.61 & 51.58 \\
            OOB-50\%       & 74.18 & 51.72 \\
            OOB-20\%       & 75.29 & 52.30 \\
            \textbf{Full Bank} & \textbf{76.01} & \textbf{52.94} \\
            \bottomrule
        \end{tabular}
    \end{minipage}
\end{figure}

\begin{table}[t]
    \centering
    \scriptsize
    \footnotesize
    \setlength{\tabcolsep}{4pt}
    \renewcommand\arraystretch{0.8}
    \caption{\textbf{Injection block placement ablation.}
Blocks 1–4 denote the first to fourth Transformer blocks (from input to output) in the visual backbone~\cite{xie2021segformer}.}
    \label{tab:audio_inject_blocks}
    \begin{tabular}{l|ccc|ccc}
        \toprule
        \multirow{2}{*}{Injected blocks}
        & \multicolumn{3}{c|}{\textit{AVS-MS3}}
        & \multicolumn{3}{c}{\textit{AVSS}} \\
        \cmidrule{2-7}
        & $\mathcal{J}\&\mathcal{F}$ & $\mathcal{J}$ & $\mathcal{F}$
        & $\mathcal{J}\&\mathcal{F}$ & $\mathcal{J}$ & $\mathcal{F}$ \\
        \midrule
        1            & 68.37 & 64.33 & 72.41 & 47.96 & 44.53 & 51.39 \\
        2            & 72.02 & 67.76 & 76.27 & 50.03 & 46.92 & 53.13 \\
        3            & 74.27 & 69.42 & 79.12 & 51.78 & 49.03 & 54.52 \\
        4            & 73.82 & 68.85 & 78.79 & 51.13 & 48.11 & 54.15 \\
        1,2          & 70.69 & 66.25 & 75.13 & 49.53 & 46.54 & 52.52 \\
        2,3          & 73.29 & 68.82 & 77.15 & 51.05 & 48.13 & 53.97 \\
        \textbf{3,4 (Ours)}   & \textbf{76.01} & \textbf{70.92} & \textbf{81.10} & \textbf{52.94} & \textbf{50.20} & \textbf{55.67} \\
        1,2,3        & 72.07 & 67.82 & 76.32 & 50.29 & 47.24 & 53.33 \\
        2,3,4        & 74.67 & 69.78 & 79.55 & 51.65 & 48.52 & 54.77 \\
        1,2,3,4      & 72.57 & 68.48 & 76.65 & 51.07 & 48.21 & 53.93 \\

        \bottomrule
    \end{tabular}
\end{table}

\begin{table}[!h]
    \centering
    \scriptsize
    \setlength{\tabcolsep}{1pt}
    \caption{\textbf{Backbone ablation.} DDAVS consistently boosts performance across all visual/audio encoder combinations.}
    \label{tab:backbone_ablation}
    \begin{tabular}{l|l|ccc|ccc|ccc}
    \toprule
    Visual Backbone & Audio Backbone & \multicolumn{3}{c|}{AVS-S4} 
                    & \multicolumn{3}{c|}{AVS-MS3} 
                    & \multicolumn{3}{c}{AVSS} \\
    \cmidrule{3-11}
                    & 
                    & $\mathcal{J}\&\mathcal{F}$ & $\mathcal{J}$ & $\mathcal{F}$ 
                    & $\mathcal{J}\&\mathcal{F}$ & $\mathcal{J}$ & $\mathcal{F}$ 
                    & $\mathcal{J}\&\mathcal{F}$ & $\mathcal{J}$ & $\mathcal{F}$ \\
    \midrule
    PVTv2-B5~\cite{wang2022pvt} & VGGish~\cite{hershey2017cnn} 
                & 91.35 & 89.36 & 93.34 
                & 73.60 & 68.77 & 78.43 
                & 47.55 & 44.11 & 50.98 \\
    PVTv2-B5~\cite{wang2022pvt} & HTSAT 
                & 91.18 & 89.12 & 93.23 
                & 74.69 & 69.49 & 79.88 
                & 48.62 & 45.53 & 51.71 \\
    MiT-B5 & VGGish~\cite{hershey2017cnn} 
                & 92.34 & 90.56 & 94.11 
                & 74.45 & 69.34 & 79.56 
                & 52.15 & 49.23 & 55.07 \\
    \textbf{MiT-B5 (Ours)} & \textbf{HTSAT (Ours)} 
                & \textbf{92.43} & \textbf{90.61} & \textbf{94.24} 
                & \textbf{76.01} & \textbf{70.92} & \textbf{81.10} 
                & \textbf{52.94} & \textbf{50.20} & \textbf{55.67} \\
    \bottomrule
    \end{tabular}
\end{table} 

\subsection{Ablation Study}

\noindent \textbf{Component Ablation.}
We evaluate DDAVS components quantitatively (\cref{tab:combined_ablation}) and qualitatively (\cref{fig:component-ablation}). 
The baseline only contains encoders, transformer blocks and the segmentation deocder.
Adding AQM yields moderate gains with minimal overhead, attributable to effective bank-based audio extraction. When COM is enabled, performance improvements become significantly more pronounced, confirming that contrastive learning is essential for robust audio representation without adding inference cost. AVAM alone delivers strong improvements by efficiently injecting audio cues for cross-modal fusion. Crucially, adding AQM+COM on top of AVAM further boosts performance from 73.16/51.45 to 76.01/52.94, confirming that optimizing query separability is essential even when spatial alignment is already established. The full DDAVS framework achieves optimal performance with $\mathcal{J}\&\mathcal{F}$ gains of +6.30 points on AVS-MS3 and +4.31 points on AVSS, while maintaining practical feasibility with only +2.16G FLOPs overhead over the baseline. Further efficiency details (FPS, training time, etc) are provided in the supplementary material.
Qualitative results (\cref{fig:component-ablation}) corroborate these findings: AQM sharpens source localization, COM improves audio-query robustness and reduces frame-wise instability, and AVAM refines spatial alignment. Their synergy yields precise, clean masks even in complex multi-source scenarios.

\noindent \textbf{Audio Query Quantity.}
\cref{fig:query_ablation} illustrates the effect of varying the number of audio queries $n$ on AVS-MS3 and AVSS. 
As $n$ increases from 1 to 5, $\mathcal{J}\&\mathcal{F}$ rises rapidly on both datasets, indicating that a small set of diverse queries helps capture different sounding patterns. 
When $n$ becomes larger than 5, the performance starts to decrease, suggesting that using too many queries is unnecessary in practice. 
Based on this empirical observation, we adopt a moderate value $n=5$ as the default setting of DDAVS.

\noindent \textbf{Injection Blocks in AVAM.}
To determine the optimal placement of cross-modal alignment within AVAM, we conduct a controlled ablation by freezing the visual backbone and training only the cross-attention modules, thereby isolating the effect of injection position from backbone updates. 
Quantitative results in \cref{tab:audio_inject_blocks} show that later-layer injection (\{3,4\}) consistently outperforms early fusion and other alternatives, achieving peak $\mathcal{J}\&\mathcal{F}$ performance. This configuration is adopted as our default setting. 
Attention maps in \cref{fig:attn_map} provide qualitative evidence for the ablation results, revealing that deeper injection enables precise instance-level focusing rather than diffuse background responses.

\noindent \textbf{Out-of-Bank Robustness.}
We evaluate robustness by progressively removing $x\%$ of class prototypes from the memory bank. 
As \cref{tab:oob} shows, DDAVS degrades gracefully: even with \emph{all} prototypes absent (OOB-100\%), it consistently outperforms the baseline while retaining a clear margin below the full-bank setting. 
This confirms prototype anchoring provides stable grounding for unseen sounds.

\noindent \textbf{Backbone Variants.}
\label{sec:backbone_variants}
\cref{tab:backbone_ablation} evaluates DDAVS across diverse visual/audio backbone combinations. 
HTSAT consistently outperforms VGGish~\cite{hershey2017cnn}, confirming its superior acoustic representation capability. 
MiT-B5 surpasses PVTv2-B5~\cite{wang2022pvt} across all benchmarks, validating its effectiveness for dense prediction tasks. 
Critically, DDAVS improves every configuration, proving the framework's architecture-agnostic design and strong generalization across backbone choices.

\begin{figure}[t]
    \centering
    \includegraphics[width=1\linewidth]{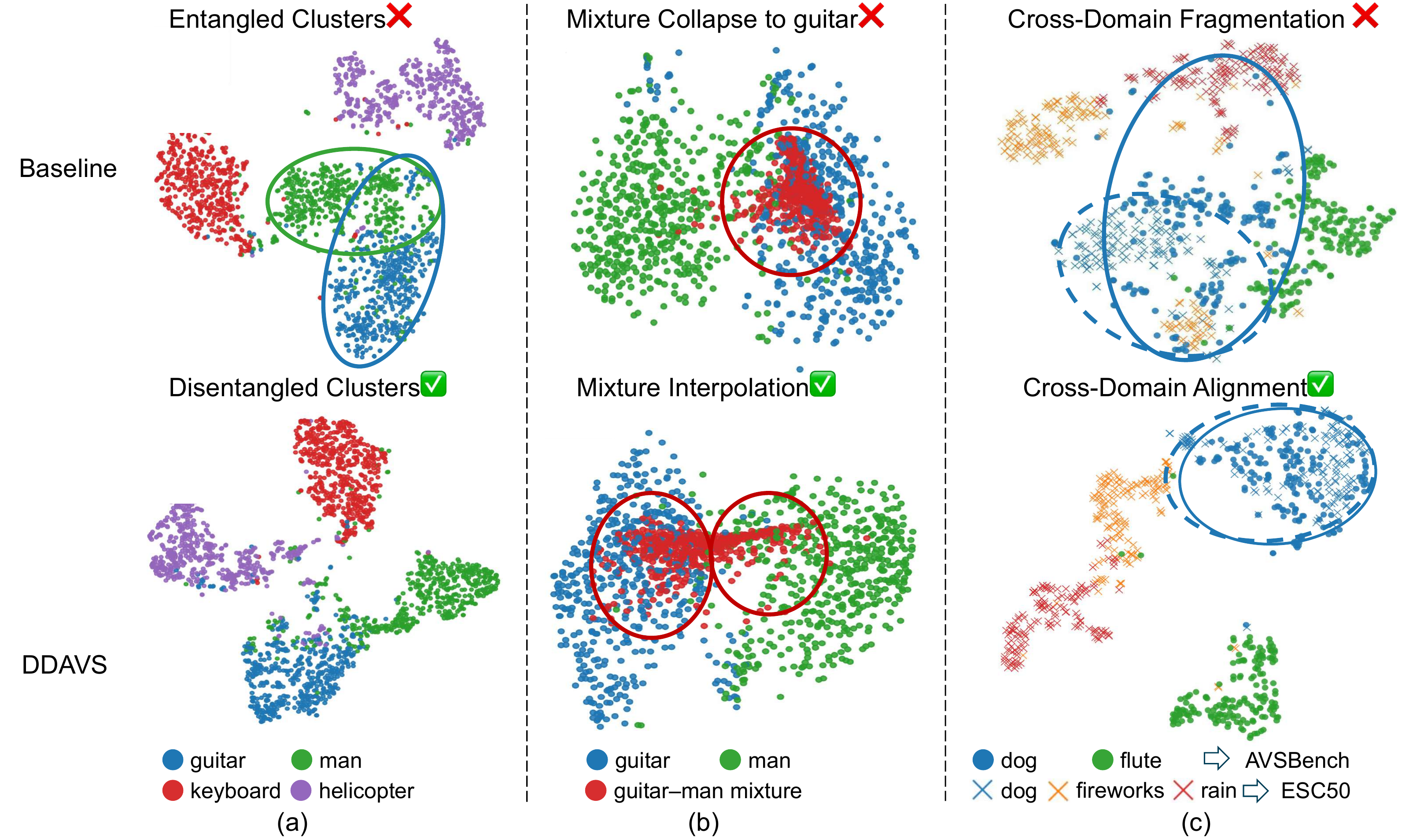}
    \caption{\textbf{t-SNE visualization of audio representations.} 
    Left and middle columns illustrate multi-source disentanglement on AVSBench. Right column evaluates cross-domain consistency for the shared class \textit{dog}.}
    \label{fig:tsne}
    \vspace{-5pt}
\end{figure}

\subsection{Representation Analysis}
\label{sec:representation_analysis}
\cref{fig:tsne} presents t-SNE visualizations that validate the quality of audio representations in terms of multi-source disentanglement (left and middle) and cross-domain consistency (right).

For \textbf{multi-source disentanglement}, \cref{fig:tsne}(a) shows single-source categories (\textit{guitar}, \textit{man}, \textit{keyboard}, \textit{helicopter}). The baseline exhibits entangled clusters with substantial overlap, while DDAVS achieves well-separated, distinct clusters. In \cref{fig:tsne}(b), the baseline suffers mixture collapse, where the \textit{guitar-man} mixture collapses into the \textit{guitar} cluster. DDAVS places mixtures along smooth interpolations between their source components, explicitly representing all constituent sources without collapsing to any single one.

For \textbf{cross-domain consistency}, we evaluate \textit{dog} embeddings across AVSBench and ESC-50~\cite{piczak2015dataset} (a dataset of 2,000 environmental audio clips). The baseline exhibits \textbf{cross-domain fragmentation}, with \textit{dog} embeddings dispersed across the embedding space according to dataset origin, undermining intra-class cohesion. In contrast, DDAVS achieves \textbf{cross-domain alignment}, consolidating all \textit{dog} embeddings into a single compact cluster that remains distinctly separated from other semantic classes (e.g., \textit{flute}, \textit{fireworks}).

These results show that DDAVS’s disentangled mixture representations and domain-invariant embeddings underlie its consistent performance gains.

\section{Discussion}
\noindent \textbf{Conclusion.}
In this work, we presented DDAVS, a novel audio–visual segmentation framework that explicitly addresses the challenges posed by multi-source mixtures and audio–visual misalignment. By introducing a prototype-guided Audio Query Module (AQM), a waveform-level Contrastive Optimization Module (COM), and a delayed bidirectional Audio–Visual Alignment Module (AVAM), our method improves semantic separation, preserves weak or mixed audio cues, and achieves more reliable cross-modal alignment. We further validate the effectiveness of this disentanglement–alignment paradigm through comprehensive experiments on AVSBench and VPO, where DDAVS establishes state-of-the-art performance across single-source, multi-source, and semantic-source settings. These results demonstrate the value of structured audio semantics and robust alignment strategies for advancing audio–visual segmentation.

\noindent \textbf{Limitations and Future Work.}
Current validation is limited to benchmark scenarios with well-defined misalignment ranges and curated sound categories. In future work, we plan to extend this paradigm to open-domain video analysis and streaming audio applications, paving the way for real-time, scalable, and broadly generalizable audio-visual perception systems.
\section*{Acknowledgement}
This work was supported in part by the National Natural Science Foundation of China under Grant 62572270, and in part by the Guangdong Natural Science Funds for Distinguished Young Scholar (No. 2025B1515020012).

\clearpage
% \mbox{}Page \thepage\ of the manuscript.
% \section*{Acknowledgements}
% Please insert your acknowledgments here.

\bibliographystyle{splncs04}
\bibliography{main}

\end{document}